\documentclass[conference]{IEEEtran}

\usepackage{xcolor}
\usepackage{graphicx}

\usepackage{amsthm}
\usepackage{amsmath}
\usepackage{amssymb}
\theoremstyle{definition}
\newtheorem{definition}{Definition}[section]
\usepackage{amsthm}

\usepackage{mathtools}
\usepackage[T1]{fontenc}

\begin{document}

\title{The Blockchain Game: 
\\Synthesis of Byzantine Systems and Nash Equilibria}

\author{\IEEEauthorblockN{Dongfang Zhao}
\IEEEauthorblockA{University of Nevada}
}

\maketitle

\begin{abstract}
This position paper presents a synthesis viewpoint of blockchains from two orthogonal perspectives: fault-tolerant distributed systems and game theory.
Specifically, we formulate a new game-theoretical problem in the context of blockchains and sketch a closed-form Nash equilibrium to the problem.
\end{abstract}

\section{Introduction}

Blockchains have drawn much research interest, way beyond its first realization, Bitcoin~\cite{bitcoin}, a cryptocurrency application built upon blockchains.
From system perspectives, various facets, especially performance and scalability, have been intensively studied by multiple computer systems communities including but not limited to: 
computer security~\cite{jlind_sosp19}, distributed systems~\cite{jwang_nsdi19}, and database systems~\cite{chainprov_vldb19}.
Works on the theoretical foundation of blockchains are, however,
comparatively limited, and mostly in the cryptocurrency context~\cite{ieyal_oakland15,rpass_disc17,itsabary_ccs18},
usually in a permissionless setup where nodes are free to join or leave the blockchain network.
In permissioned blockchains such as Hyperledger Fabric~\cite{hyperledger_eurosys18}, where Practical Byzantine Fault-Tolerance~\cite{castro_osdi99} (PBFT) is the \textit{de facto} consensus protocol, 
much work focused on PBFT and its variants without in-depth reasoning on the node's (or, user's) \textit{rationality}---analyses simply assume that a node is either \textit{faulty} or \textit{non-faulty}.
Admittedly, the emphasis of reasoning about a node's rationality is historically an area in multi-agent systems and game theory.

This position paper envisions and advocates that a global-scale, likely utility-incentive, blockchain system might be better understood from a holistic view of multiple perspectives as its theoretical foundation.
As a starting point, we take a hybrid approach of both distributed computing and game theory to study blockchains.
The inner-connection between distributed computing and game theory dates back to 2011~\cite{iabraham_sigact11},
which reviewed important commonality between, and more importantly complimentary research methodologies shared by two communities,
both of which parallelly concentrated on distributed irrational-machine systems and utility-maximizing multi-agent systems, respectively.
We will articulate the models of our approach in~\S\ref{sec:model},
formulate the \textit{Blockchain Game} in~\S\ref{sec:game} problem,
sketch one possible Nash equilibrium for the problem and discuss open questions in~\S\ref{sec:discuss}.

\section{Models}
\label{sec:model}

\begin{figure}[!th]

  \centering
  \includegraphics[width=85mm]{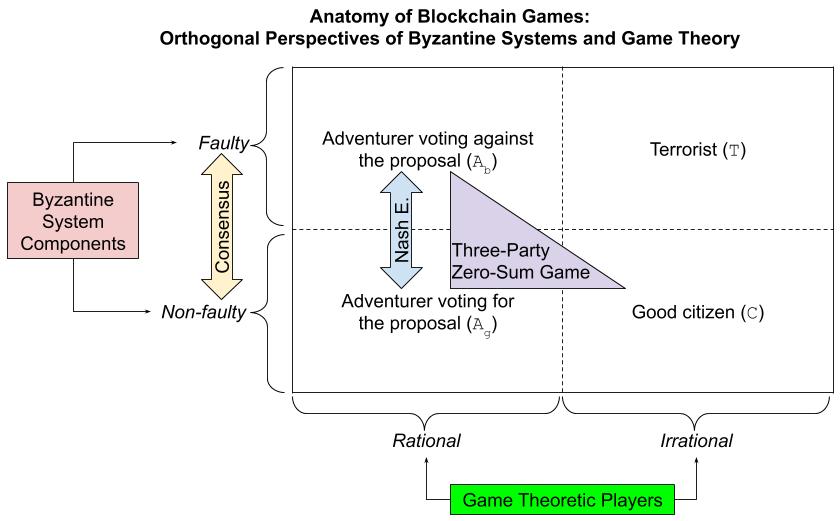}
  \caption{The blockchain game can be viewed from two orthogonal perspectives: Byzantine distributed systems and game theory.
  }
    \label{fig:diagram}  
\end{figure}

We illustrate the \textit{Blockchain Game} in Figure~\ref{fig:diagram}.
There are three type of nodes\footnote{Also called players (game theory), users (databases), or participants (distributed systems), depending on various contexts.} involved in a blockchain game:
a good citizen (\texttt{C}) who always votes for the proposal,
a terrorist (\texttt{T}) who always votes against the proposal,
and an adventurer (\texttt{A}) who makes her decision to maximize the utility (either voting for ($\mathtt{A_g}$) or against ($\mathtt{A_b}$) the proposed value, i.e., $\mathtt{A} = \mathtt{A_g} \cup \mathtt{A_b}$).
A proposal could be a proposed value in a permissioned blockchain or a newly-mined block to be verified by a permissionless blockchain.

We use $\mathcal{N} = \mathtt{C} \cup \mathtt{T} \cup \mathtt{A}$ to denote the entire set of nodes, and use $\mathbf{n} = |\mathcal{N}|$ the cardinality of the node set.
We will use $\mathbf{t} = |\mathtt{T}|$ to denote the cardinality of the terrorist set $\mathtt{T}$,
$\mathbf{c} = |\mathtt{C}|$ to denote the cardinality of the citizen set $\mathtt{C}$,
and $\mathbf{g} = |\mathtt{A_g|}$ and $\mathbf{b} = |\mathtt{A_b|}$
for the two subsets' cardinalities, respectively. 

\section{The Blockchain Game}
\label{sec:game}

\subsection{Normal Game Strategies}
\label{subsec:game_strategy}

In the real world, a distributed system, including blockchains, usually adopts a \textit{timeout} mechanism that sets some unresponsive nodes to a default value, such as \textit{null} or \textit{nil}.
From a pure system point of view, an \textit{abstain} vote is not different than a \textit{no} vote---the system takes a very conservative position in interpreting the responses.
In the case of blockchains, therefore, the consensus protocol interprets an \textit{abstain} vote as a \textit{no} vote.
Formally, we have two pure strategies for all the $\mathbf{n}$ nodes in the blockchain: 
$\mathcal{S}_i = \{\mathtt{Yes}, \mathtt{No}\}$, $0 \le i < \mathbf{n}$.

In blockchains, it takes a simple quorum mechanism to move forward.
In permissionless blockchains (Bitcoin~\cite{bitcoin}, Ethereum~\cite{ethereum}), the longest list of transactions will overwrite the shorter ones and the system is considered stable as long as no more than 51\% nodes are controlled or compromised (i.e., $\mathbf{g} + \mathbf{c} > \mathbf{b} + \mathbf{t}$).
In permissioned blockchains (Hyperledger Fabric~\cite{hyperledger_eurosys18} and other variants based on PBFT~\cite{castro_osdi99}), 
non-faculty nodes need to outnumber the faulty nodes by at least 200\%.
The following discussion assumes the blockchain is permissionless for the sake of space.

If the system is not compromised, the non-faulty nodes will continue to work on ``agreeing'' on the next proposed value and get rewarded by a transaction fee, and the faulty nodes might be forced to leave the network;
Otherwise, the faulty nodes overturn the existing network and collect a high reward, leaving the ``honest'' nodes' work worthless.
Note that we neglect the cost (e.g., electricity, hardware procurement, space rental) for mining blocks or attacking the network.
Formally, the payoff function for a non-faulty node $u_n: \mathtt{A_g} \cup \mathtt{C} \mapsto \mathbb{R}$ is defined as follows:
\begin{equation}
\label{eq:payoff1}
    u_n \triangleq 
\begin{cases}
    \frac{\displaystyle p_n}{\displaystyle \mathbf{g} + \mathbf{c}},            & \text{if } \mathbf{g} + \mathbf{c} > \mathbf{b} + \mathbf{t}\\
    0,              & \text{otherwise}
\end{cases}
\end{equation}
where $p_n$ denotes the overall payoff of a winning ``honest'' nodes and $n$ denotes that winning non-faulty node, i.e., $n \in \mathtt{A_g} \cup \mathtt{C}$.
Statistically, each ``honest'' node receives $\frac{\displaystyle p_n}{\displaystyle \mathbf{g} + \mathbf{c}}$ payoff.
Similarly, we have the payoff for a faulty node (but not a terrorist) as follows:
\begin{equation}
\label{eq:payoff2}
    u_f \triangleq 
\begin{cases}
    \frac{\displaystyle p_f}{\displaystyle \mathbf{b}},     
    & \text{if } \mathbf{g} + \mathbf{c} \le \mathbf{b} + \mathbf{t}\\
    0,              & \text{otherwise}
\end{cases}
\end{equation}
where $p_f$ denotes the overall payoff of all ``deviating'' nodes and $f$ denotes a faulty node, i.e., $f \in \mathtt{A_b}$.

\subsection{Zero-Sum among Non-Terrorist Nodes}
An important observation is that all the non-terrorist nodes constitute a zero-sum game.
That is, the Byzantine nodes, or terrorists, are disinterested in the expense or utility incurred in the game---the only objective is to sabotage the network.
As a consequence, we have the following invariant:
\begin{equation}
\label{eq:zerosum}
    0 \equiv \mathbf{c} \cdot p_n + \mathbf{g} \cdot p_n - \mathbf{b} \cdot p_f
\end{equation}
where the first term indicates the payoff for good citizen,
the second term indicates the payoff for adventurers who vote for the proposal,
the third term indicates the payoff for adventurers who vote against the proposal.

\subsection{Consensus Protocols}
\label{subsec:game_consensus}

There are rich literature in handling \textit{arbitrary} nodes in the distributed system community. 
In the context of blockchains,
there are two main categories of \textit{consensus protocols} by which the participating reach an agreement:
(i) Proof-of-Work (PoW) and its variants for permissionless blockchains; (ii) PBFT for permissioned blockchains.
Again, we will focus on permissionless blockchains in this paper.

Permissionless blockchains require that the number of non-faulty nodes is strictly larger than that of faulty nodes. 
Therefore, we set the difference as one between the two cliques as the borderline case in the following discussion:
\begin{equation}
\label{eq:consensus_public}
    \mathbf{c} + \mathbf{g} \geq 1 + \mathbf{t} + \mathbf{b} 
\end{equation}

\subsection{Blockchain Game}

With all the aforementioned preliminaries, we are ready to define a general blockchain system and the associated game.

\begin{definition}[Blockchain System]
\label{def:blockchain_system}
A blockchain system $\mathfrak{B}$ is represented as a tuple $\mathfrak{B} = \langle \mathcal{N}, \mathcal{S}, \mathcal{P} \rangle$,
where $\mathcal{N}$ denotes the set of all nodes (or, players),
$\mathcal{S}$ denotes the set of strategies and associated payoffs available for the nodes,
and $\mathcal{P}$ denotes the consensus protocol among the nodes.
\end{definition}

\begin{definition}[Blockchain Game]
In a blockchain system $\mathfrak{B}$, each \textit{adventurer} node in $\mathcal{N}$ of a blockchain maximizes its utility according to $\mathcal{S}$ without violating $\mathcal{P}$.
\end{definition}

Essentially, the \textit{rational} players would choose between being part of either $\mathtt{A_g}$ or $\mathtt{A_b}$, exclusively,
such that her utility is maximized under multiple constraints, i.e., equations~\ref{eq:payoff1}--\ref{eq:consensus_public}.

\section{Discussions}
\label{sec:discuss}

Since the \textit{Blockchain Game} exhibits a finite number of players and strategies,
there must exist at least one Nash equilibrium.
Obviously, the solution to the above equations represents one such equilibrium.
Due to limited space, we simply give the closed-form solution without numerical analysis:
\begin{equation}
\label{eq:equil_result}    
\begin{cases} 
\mathbf{g^*} = \frac{\displaystyle 1 + \mathbf{t}}{\displaystyle 1 - \gamma} - \mathbf{c} \\
\mathbf{b^*} = \frac{\displaystyle \gamma\cdot(1 + \mathbf{t})}{\displaystyle 1 - \gamma}
\end{cases}
\end{equation}
where $\gamma = \frac{\displaystyle p_n}{\displaystyle p_f}$.
Note that in the real world, $p_f$ is, usually, significantly larger than $p_n$, implying that $0 < \gamma << 1$.
We call this variable \textit{reciprocal risk factor} (RRF),
indicating the payoff ratio of a compliant action over a deviating action.

It should be clear that the results are for \textit{permissionless blockchains} only,
although a similar one can be obtained for \textit{permissioned blockchains} as well.
It should also be noted that the discussion so far is limited to a normal game, 
which is played by the nodes only \textit{once}.
We leave \textit{extensive-form game} that counts times as an open question and our future work.

\bibliographystyle{IEEEtranS}
\bibliography{NDSS2020_abstract}

\begin{thebibliography}{10}
\providecommand{\url}[1]{#1}
\csname url@samestyle\endcsname
\providecommand{\newblock}{\relax}
\providecommand{\bibinfo}[2]{#2}
\providecommand{\BIBentrySTDinterwordspacing}{\spaceskip=0pt\relax}
\providecommand{\BIBentryALTinterwordstretchfactor}{4}
\providecommand{\BIBentryALTinterwordspacing}{\spaceskip=\fontdimen2\font plus
\BIBentryALTinterwordstretchfactor\fontdimen3\font minus
  \fontdimen4\font\relax}
\providecommand{\BIBforeignlanguage}[2]{{%
\expandafter\ifx\csname l@#1\endcsname\relax
\typeout{** WARNING: IEEEtranS.bst: No hyphenation pattern has been}%
\typeout{** loaded for the language `#1'. Using the pattern for}%
\typeout{** the default language instead.}%
\else
\language=\csname l@#1\endcsname
\fi
#2}}
\providecommand{\BIBdecl}{\relax}
\BIBdecl

\bibitem{iabraham_sigact11}
I.~Abraham, L.~Alvisi, and J.~Y. Halpern, ``Distributed computing meets game
  theory: Combining insights from two fields,'' \emph{SIGACT News}, 2011.

\bibitem{hyperledger_eurosys18}
E.~Androulaki \emph{et~al.}, ``Hyperledger fabric: A distributed operating
  system for permissioned blockchains,'' in \emph{EuroSys}, 2018.

\bibitem{bitcoin}
{Bitcoin}, ``\url{ https://bitcoin.org/bitcoin.pdf},'' Accessed 2018.

\bibitem{castro_osdi99}
M.~Castro and B.~Liskov, ``Practical byzantine fault tolerance,'' in
  \emph{Operating Systems Design and Implementation (OSDI)}, 1999.

\bibitem{ethereum}
{Ethereum}, ``\url{ https://www.ethereum.org/},'' Accessed 2018.

\bibitem{ieyal_oakland15}
I.~{Eyal}, ``The miner's dilemma,'' in \emph{IEEE Symposium on Security and
  Privacy}, 2015.

\bibitem{jlind_sosp19}
J.~Lind \emph{et~al.}, ``Teechain: A secure payment network with asynchronous
  blockchain access,'' in \emph{ACM Symposium on Operating Systems Principles
  (SOSP)}, 2019.

\bibitem{rpass_disc17}
R.~Pass and E.~Shi, ``{Hybrid Consensus: Efficient Consensus in the
  Permissionless Model},'' in \emph{31st International Symposium on Distributed
  Computing (DISC)}, 2017.

\bibitem{chainprov_vldb19}
P.~Ruan \emph{et~al.}, ``Fine-grained, secure and efficient data provenance on
  blockchain systems,'' \emph{Proc. VLDB Endow.}, 2019.

\bibitem{itsabary_ccs18}
I.~Tsabary and I.~Eyal, ``The gap game,'' in \emph{ACM SIGSAC Conference on
  Computer and Communications Security (CCS)}, 2018.

\bibitem{jwang_nsdi19}
J.~Wang and H.~Wang, ``Monoxide: Scale out blockchains with asynchronous
  consensus zones,'' in \emph{{USENIX} Symposium on Networked Systems Design
  and Implementation (NSDI)}, 2019.

\end{thebibliography}

\end{document}